# The AI Pyramid

## A Conceptual Framework for Workforce Capability in the Age of AI

**Alok Khatri [1,2] and Bishesh Khanal [1,2]**

1. NAAMII, Nepal 2. Tangible Careers

## Abstract

Artificial intelligence (AI) represents a qualitative shift in technological change by extending cognitive labor itself rather than merely automating routine tasks. Recent evidence shows that generative AI disproportionately affects highly educated, white-collar work, challenging existing assumptions about workforce vulnerability and rendering traditional approaches to digital or AI literacy insufficient. This paper introduces the concept of AI Nativity, the capacity to integrate AI fluidly into everyday reasoning, problem solving, and decision-making, and proposes the AI Pyramid, a conceptual framework for organizing human capability in an AI-mediated economy. The framework distinguishes three interdependent capability layers: AI Native capability as a universal baseline for participation in AI-augmented environments; AI Foundation capability for building, integrating, and sustaining AI-enabled systems; and AI Deep capability for advancing frontier AI knowledge and applications. Crucially, the pyramid is not a career ladder but a system-level distribution of capabilities required at scale. Building on this structure, the paper argues that effective AI workforce development requires treating capability formation as infrastructure rather than episodic training, centered on problem-based learning embedded in work contexts and supported by dynamic skill ontologies and competency-based measurement. The framework has implications for organizations, education systems, and governments seeking to align learning, measurement, and policy with the evolving demands of AI-mediated work, while addressing productivity, resilience, and inequality at societal scale.

## Introduction

Technological change has historically amplified distinct dimensions of human capability. Industrial machinery extended physical strength, digital computation accelerated numerical processing, and the internet enabled unprecedented coordination of information. Artificial intelligence (AI) represents a qualitatively different shift: it extends cognitive labor itself. Contemporary AI systems now perform tasks like writing, summarizing, pattern recognition, communication, ideation, and prediction, forms of reasoning that were, until recently, exclusively human domains.

Prior research on technological change suggests that automation historically concentrated on routine forms of work. In contrast, recent evidence shows that generative AI disproportionately affects highly educated, white-collar occupations engaged in non-routine cognitive tasks (Felten et al., 2023). This reversal challenges existing assumptions about workforce vulnerability and demands a rethinking of how capability development is organized.

Existing frameworks, "digital literacy," "AI awareness," or basic tool-use training, fail to accommodate this transformation. These models treat AI as an external instrument that individuals must understand and operate. In practice, however, working with contemporary AI systems requires people to negotiate, evaluate, and co-produce outputs with agents that function less like passive tools and more like cognitive collaborators.

To capture this emergent mode of interaction, this paper introduces the concept of **AI Nativity**. AI Nativity does not imply technical mastery or the ability to build AI systems. Rather, it describes the capacity to integrate AI fluidly into one's thinking, problem-solving routines, and everyday cognitive workflows, treating AI as part of the environment in which reasoning occurs. Building on this orientation, the paper proposes the **AI Pyramid**, a framework that both categorizes the capability segments societies require and guides how those capabilities can be systematically developed through training and infrastructure in an AI-mediated economy.

## From AI Literacy to AI Nativity

AI literacy typically refers to a foundational understanding of Artificial Intelligence. Long and Magerko (2020) provide a comprehensive academic framework, identifying 17 competencies across five themes for AI: what it is, what it can do, how it works, how it should be used, and how people perceive it. While necessary, literacy does not capture the deeper behavioral and cognitive adjustments required to work with AI systems that now generate content, propose solutions, reason through problems, and adapt dynamically to feedback. Literacy prepares individuals to recognize AI; it does not prepare them to think and live with AI.

AI nativity describes a more embedded and intuitive orientation. Drawing on situated learning theory (Lave & Wenger, 1991), AI-native individuals treat AI systems as collaborators within their communities of practice rather than external tools. Recent empirical evidence demonstrates the power of this integration: customer support agents using generative AI assistance increased productivity by 15% on average, with novice workers experiencing 30% improvement, suggesting AI disseminates expert practices to less experienced workers (Brynjolfsson et al., 2025).

Crucially, the shift from literacy to nativity is not a linear progression but a qualitative transformation, analogous to the difference between studying a foreign language and achieving native fluency. AI nativity reflects the capacity to integrate AI into the cognitive ecology of work, using AI as part of one's reasoning environment rather than as an external resource.

## The AI Pyramid Framework

Workforce planning in the AI era requires clarity about what different population segments need to know and do. This paper proposes the **AI Pyramid**, a framework that distinguishes three levels of capability required for an AI-ready workforce. The framework builds on Lepak and Snell's (1999) Human Resource

Architecture theory, which argues that organizations should differentiate HR investments based on the strategic value and uniqueness of different human capital segments.

The pyramid's shape represents population scale and functional dependency, not value hierarchy. While visual convention might suggest the apex represents the "highest" achievement, our framework inverts this assumption: the base is the most critical layer. Just as ancient pyramids required massive foundations to support narrow peaks, AI-enabled societies require broad AI Native capability to support smaller populations of Foundation builders and Deep researchers. Without a stable base, nothing above it can function. The pyramid shows what is required for the whole system to work, not a ladder individuals should climb.

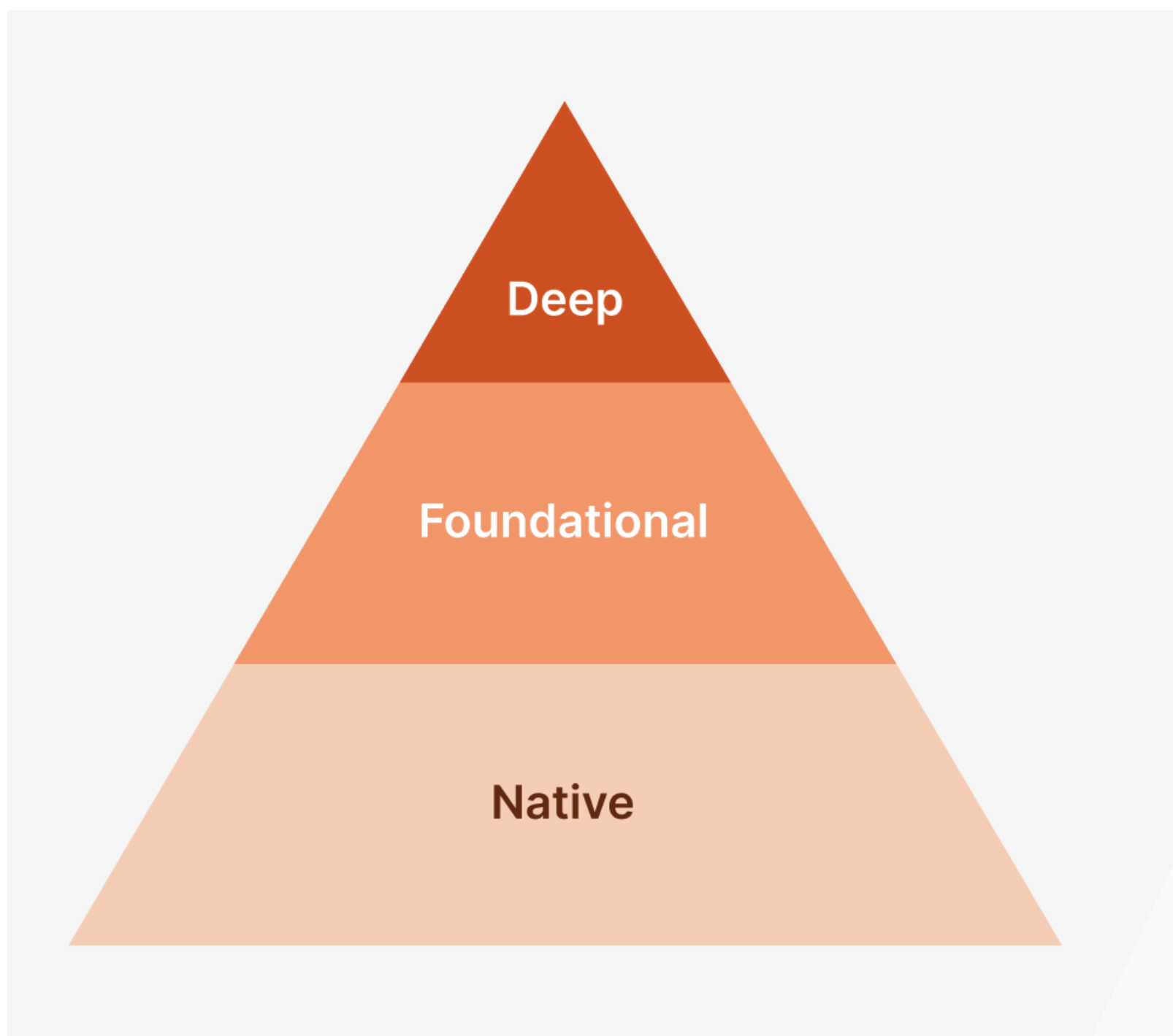

**AI Native Capability**

AI Native capability forms the base of the pyramid. It encompasses the broad population that works and reasons in AI‑mediated environments but does not design, train, or engineer AI systems. AI nativity does not refer to technical proficiency, domain expertise, or superior cognitive ability in isolation. Rather, it describes a behavioral orientation: the ability to routinely frame problems in ways AI systems can meaningfully engage, to evaluate and contextualize AI outputs critically, and to integrate machine‑generated suggestions into human judgment and decision making as part of everyday cognitive workflows. Central to this orientation is awareness of AI's limitations, potential biases, and risks of misuse, including the capacity to recognize when AI outputs should be questioned, constrained, or excluded from decision processes. This capability is observable not in what individuals know about AI, but in how they habitually incorporate, interrogate, and govern AI use in task formulation, iteration, and problem resolution across diverse contexts.

Widespread AI Native capability is essential because it shapes how teams coordinate work, how organizations make decisions, and how institutions adapt in environments where AI increasingly mediates information, analysis, and communication. Evidence from exposure research underscores the need for this capability to extend beyond technical roles. Felten et al. (2023) show that generative AI most strongly affects highly educated, highly paid white-collar occupations engaged in non-routine cognitive work, indicating that AI nativity cannot be confined to specialized users or advanced adopters. Instead, it must function as a baseline capability across knowledge work broadly, enabling consistent, responsible, and effective participation in AI-augmented organizational and societal processes. In principle, AI-native capability can be assessed through observable patterns of interaction with AI systems, including how individuals frame tasks, evaluate outputs, and exercise judgment in contexts involving uncertainty, risk, or ethical trade-offs.

**AI Foundation Capability**

AI Foundation capability constitutes the middle layer of the pyramid and refers to individuals who design, build, integrate, and maintain AI-enabled systems within organizations. This group includes engineers, technical generalists, applied data practitioners, and developers who translate organizational objectives into deployable models, tools, and workflows. AI Foundation capability is distinct from AI nativity in that it involves system construction rather than system use, but it does not require frontier research expertise. Its defining feature is the ability to operationalize AI reliably in real contexts, connecting data, models, infrastructure, and organizational processes while accounting for constraints such as performance, scalability, security, and governance.

Unlike AI Native capability, which emphasizes behavioral fluency, AI Foundation capability is expressed through what individuals can build, adapt, and sustain over time. This layer faces the most acute challenge of skill obsolescence, as the technical components underlying AI systems evolve rapidly. Deming and Noray (2020) show that the economic returns to applied STEM degrees decline sharply within the first decade of working life due to rapid turnover in skill content. As a result, Foundation-level capability cannot be secured through static credentials or one-time training. It requires continuous adaptive development, problem-based learning anchored in real implementation challenges, and ongoing exposure to evolving tools, architectures, and deployment practices. Capability at this level is therefore best assessed through demonstrated system-building and integration outcomes rather than formal qualifications or one-time credential acquisition alone.

**AI Deep Capability**

AI Deep capability occupies the apex of the pyramid and refers to individuals who advance the frontiers of artificial intelligence itself. This group includes researchers, scientists, and engineers who develop new model architectures, learning algorithms, training methods, and theoretical insights, as well as those who apply frontier AI techniques to generate breakthroughs across specific domains such as health, education, science, energy, and public policy. AI Deep capability extends beyond system implementation toward the creation of new capabilities, expanding what AI systems can do and how they can be applied across sectors.

Unlike AI Native and AI Foundation capabilities, AI Deep capability is not required within every organization, nor is it optimal for all institutions to cultivate it internally. Instead, this layer operates primarily at the level of research ecosystems, national innovation systems, and global knowledge networks. Its impact is realized through spillovers: advances generated by a relatively small number of deep specialists diffuse outward, shaping the tools, platforms, and methods used by Foundation-level builders and, ultimately, AI-native users. As a result, the strategic importance of AI Deep capability lies not in organizational ubiquity but in societal and national capacity to contribute to, access, and govern frontier AI development. Capability at this level is inseparable from sustained investment in research infrastructure, computational resources, and collaborative environments where theoretical and applied advances co-evolve.

## Capability Distribution and the Infrastructure Challenge

The pyramid is not intended as a career ladder or a normative progression that individuals are expected to climb. Its structure reflects differences in scale, function, and dependency, not relative status or achievement. AI-enabled societies do not require all individuals or organizations to move toward higher layers of specialization; rather, they depend on a stable distribution of capabilities across layers. Broad AI Native capability enables effective participation in AI-mediated environments, Foundation capability enables systems to be built and sustained, and Deep capability generates frontier advances whose benefits diffuse outward. The pyramid therefore describes how capabilities must be distributed across a system, not how individuals should advance within it.

Workforce planning that overemphasizes one layer at the expense of others risks creating systemically imbalanced organizations and societies, with sophisticated research capacity but limited deployment expertise, or widespread tool adoption without the technical infrastructure to sustain it.

### The Distribution Challenge

The distribution of individuals across these three layers directly determines what communities, organizations, and governments can achieve with AI. Global evidence reveals significant geographic disparities: while approximately 60% of employment is exposed to AI in advanced economies, only 26% faces exposure in low-income countries due to different labor market structures (Cazzaniga et al., 2024). Importantly, this exposure encompasses both opportunities for productivity gains and risks of displacement, with advanced economies better positioned to capture the benefits.

More significantly, when large segments of a population become AI-native, the cognitive environment itself transforms (Clark, 2025).. Conversations shift as more participants can reason fluently with AI. The range of feasible solutions expands. Organizational learning accelerates. This represents not merely productivity gains but a fundamental change in institutional adaptability and collective problem-solving capacity.

### Becoming AI Native

AI-native capabilities comprise a cluster of competencies that enable individuals to participate effectively in AI-mediated cognitive work. Building on Long and Magerko’s (2020) AI literacy framework and extending it toward application, these capabilities center on five core skills: formulating tasks in alignment with how models interpret and reason; evaluating the reliability and limits of AI outputs; orchestrating AI‑enabled workflows; integrating machine‑generated suggestions into human judgment and decision‑making; and exercising responsible AI awareness, including recognizing potential biases, misuse, and ethical risks, and knowing when AI outputs should be constrained, contextualized, or excluded from use.

In practice, these competencies manifest through four interconnected skill areas: **Prompting** (the ability to articulate problems and constraints in ways AI systems can act upon consistently), **AI automation** (the design of workflows where AI performs structured or repeatable tasks), **AI-enabled problem solving** (the capacity to combine human judgment with machine reasoning to generate solutions), and **AI literacy** (foundational understanding of how AI systems behave, where they perform well, where they fail, and risks involved on potential misuse).

These skills are not static. They evolve as AI systems expand in reasoning depth, memory, and interaction capabilities. What distinguishes the current phase of AI adoption is that AI-native skills now constitute the baseline layer of capability the entire workforce requires, not a specialized technical skillset, but the baseline for productive participation in AI-enabled environments.

**From Programs to Infrastructure**

Historically, capability building operated as discrete programs, training interventions designed to transfer stable skillsets. Individuals attended courses, acquired certifications, and entered roles where those skills remained relevant for extended periods. This model assumed that knowledge, once learned, would retain its value.

AI fundamentally disrupts this model. The skills required to work effectively with AI, prompt engineering, context management, workflow orchestration, evolve continuously. What constituted effective practice months ago may be insufficient today. This creates a structural problem: if the whole population needs to be AI-native, and multiple sectors are experiencing rapid transformation, traditional program-based training cannot keep pace.

The answer requires treating capability building as infrastructure rather than isolated programs. Yet global infrastructure for this remains incomplete. Most nations lack a coherent approach to systematic AI capability development, creating a critical gap between the pace of technological change and the pace of workforce adaptation.

Treating capability development as infrastructure requires specifying the institutional systems that sustain it. In the context of AI, this infrastructure comprises at least three interdependent components. **Measurement infrastructure** makes AI-related capabilities visible by defining, assessing, and updating what it means to be AI‑native, Foundation‑level, or Deep‑level across roles and sectors. **Learning infrastructure** provides continuous, problem‑embedded pathways through which individuals acquire and adapt these capabilities as AI systems evolve, rather than relying on episodic training interventions. **Credentialing and verification infrastructure** enables capabilities to be recognized, transferred, and trusted across organizations and labor markets, reducing reliance on static credentials that quickly lose relevance. Together, these systems allow societies to move from ad hoc training efforts toward durable, scalable capacity building aligned with the layered requirements of the AI Pyramid.

## Problem-Based Learning as a Mechanism for Capability Development

As AI capability development shifts from episodic programs to continuous infrastructure, learning mechanisms must be embedded within systems that support ongoing adaptation. Problem-based learning (PBL) serves as a core mechanism within this learning infrastructure, precisely because it aligns skill development with the real problems individuals and organizations must solve in AI-mediated environments. AI-related skills are not abstract technical knowledge but contextual practices intertwined with specific workflows and roles. PBL offers a more appropriate pedagogical approach precisely because it adapts to the problems learners actually need to solve.

The evidence base for PBL has matured substantially. Strobel and van Barneveld's (2009) meta-synthesis of PBL meta-analyses concludes that PBL is superior to traditional instruction for long-term retention, skill development, and satisfaction (both student and teacher), though traditional instruction shows advantages for short-term retention on standardized board examinations. More granularly, Gijbels et al.'s (2005) analysis finds the strongest positive effects at the level of "understanding principles that link concepts", precisely the type of understanding required for AI-augmented work.

PBL inverts the conventional training sequence. Rather than beginning with AI concepts and then seeking applications, it starts with real tasks or challenges individuals face in their roles. AI concepts enter when they are relevant to solving those specific problems. Learners engage in iterative cycles of experimentation and refinement, internalizing AI capabilities through application rather than abstraction.

The theoretical foundation comes from situated learning theory. Lave and Wenger's (1991) framework of "legitimate peripheral participation" describes how learners develop expertise by participating in authentic practices within communities. For AI capability development, this suggests embedding learning within actual work contexts rather than abstracting it into classroom environments.

For learning infrastructure to function at scale, however, problem-based learning must be coupled with competency-based measurement. AI-related capabilities cannot be reliably assessed through course completion or static credentials, because proficiency depends on how individuals perform across contexts and over time. A competency-based framework evaluates fine-grained capabilities, such as task formulation, judgment under uncertainty, workflow orchestration, or responsible AI use, along continua of proficiency rather than binary completion. This approach enables AI-native, Foundation, and Deep capabilities to be measured consistently despite variation in roles, sectors, and learning pathways.

At the system level, competency-based assessment requires underlying skill ontologies that define how AI-related capabilities are structured, related, and sequenced. Large-scale skill ontologies provide a shared language for mapping competencies to occupational roles, learning pathways, and organizational needs. When combined with competency-based evaluation, these ontologies allow capability development to be tracked longitudinally, compared across populations, and updated as AI systems evolve. In this way, skill ontologies function not as static taxonomies, but as dynamic infrastructure supporting continuous measurement, learning, and workforce planning.

At the AI Native level, a manager learning to use AI for strategic planning faces different challenges than an educator designing AI-augmented curricula. At the AI Foundation level, an engineer building recommendation systems confronts different problems than a data practitioner designing automated workflows. At the AI Deep level, researchers advancing model architectures work through fundamentally different challenges than those optimizing training efficiency. PBL allows each segment to develop capability through problems germane to their context, building practical fluency rather than generalized awareness.

## Building Capability Infrastructure Through the Pyramid

The AI Pyramid provides a framework for structuring capability infrastructure in the AI Native world. Rather than treating workforce development as a uniform challenge, the pyramid reveals that different population segments require fundamentally different capability-building approaches. This differentiation builds on Lepak and Snell's (1999) proposition that human capital with different strategic value and uniqueness requires distinct development investments.

Across all layers of the AI Pyramid, capability development relies on a shared measurement foundation: large‑scale, dynamic skill ontologies paired with competency‑based assessment. Because AI‑related capabilities evolve rapidly and manifest through practice rather than formal instruction, they cannot be reliably captured through static degrees or course completion. Instead, fine‑grained competencies, such as task formulation, judgment under uncertainty, system integration, or responsible AI use, must be continuously assessed within problem‑based learning environments where skills are demonstrated in context. Skill ontologies provide the structure for defining and updating these competencies, while competency‑based measurement enables progress to be tracked longitudinally across roles, sectors, and population groups. This shared infrastructure supports learning, verification, and workforce planning at scale, while allowing each layer of the pyramid to adapt assessment and development to its distinct functional requirements.

#### AI Native Level: Mass Accessibility and Behavioral Fluency

At the AI Native level, the shared competency‑based infrastructure must be adapted for scale and accessibility across the general population. Here, problem‑based learning is embedded directly into everyday work and civic contexts, enabling individuals to develop fluency in AI‑mediated reasoning through routine task performance rather than formal instruction. The relevant competencies emphasize behavioral patterns, how people frame problems for AI systems, evaluate outputs, exercise judgment under uncertainty, and recognize risks or misuse, rather than technical knowledge. As a result, learning and assessment at this level prioritize continuous, low‑friction integration into daily workflows, ensuring AI nativity functions as a baseline capability for participation in AI‑structured environments rather than a specialized qualification.

#### AI Foundation Level: Role‑Specific System‑Building Capability

At the AI Foundation level, the same competency‑based infrastructure is applied to a narrower population with distinct functional responsibilities: individuals who build, integrate, and maintain AI‑enabled systems on behalf of organizations. This includes engineers, applied data practitioners, technical generalists, and developers tasked with translating organizational objectives into deployable, governed, and sustainable AI workflows. Problem-based learning at this level centers on real implementation challenges, such as data pipeline design, model integration, workflow automation, system monitoring, and responsible deployment within organizational constraints. Competency assessment therefore emphasizes demonstrated system-building outcomes tied to specific roles and contexts, enabling continuous updating and re-verification of capability as tools and architectures evolve.

**AI Deep Level: Research Ecosystems**

At the AI Deep level, the shared measurement and learning infrastructure supports a small population whose work advances the frontiers of AI in theory and application. Problem-based learning here takes the form of direct engagement with open research problems, whether in developing new algorithms and architectures or applying frontier techniques to generate breakthroughs across domains such as science, health, education, and public policy. Unlike the lower layers of the pyramid, this capability is not required within every organization; its value emerges primarily at societal, national, and global levels through knowledge spillovers that shape downstream tools and practices. Assessment at this level therefore focuses on frontier contribution and diffusion, research outputs, applied impact, and influence on broader ecosystems, rather than standardized credentials or role-based benchmarks.

The AI Pyramid implies that developing human capability in an AI-mediated economy requires a layered but unified infrastructure, in which a shared, competency-based measurement and learning foundation is adapted, not reinvented, across AI Native, Foundation, and Deep capabilities. Societies that invest in only one layer, such as mass AI awareness without system-building capacity, or frontier research without broad adoption and governance, risk creating structural capability imbalances that limit the social and economic returns to AI. Effective workforce development therefore depends not on isolated programs, but on coherent infrastructure that aligns measurement, learning, and capability distribution across the entire pyramid.

## From Organizational Implementation to National Infrastructure

Organizations can operationalize the AI Pyramid by ensuring broad AI Native capability across their workforce while selectively accessing higher layers as required by their strategic context. The foundation requires universal adoption: everyone must become AI-native, developing fluency in AI-augmented work through problem-based learning integrated into everyday tasks. Some organizations will also cultivate or hire AI Foundation capability to build and maintain AI-enabled systems internally, while others may rely on external providers or shared platforms. AI Deep capability, by contrast, is rarely required within individual organizations and is typically accessed through partnerships, research ecosystems, or national and global innovation systems rather than developed in-house.

The empirical evidence supports this emphasis on broad AI Native capability. Brynjolfsson et al. (2025) find that AI assistance delivers the largest productivity gains to lower-skilled workers (approximately 30%), while gains for higher-skilled workers are smaller; however, the overall productivity improvement of 15% depends on widespread adoption. This suggests that organizations cannot selectively deploy

AI‑native capability to a subset of roles but must develop it broadly across the workforce to realize system‑level benefits.

However, what works at the organizational level becomes inadequate at societal scale. Society‑wide capability development faces structural challenges: skills decay faster than formal education systems can update curricula (Deming & Noray, 2020); societies lack coherent skill ontologies for AI‑related work; and pre‑employment populations often lack access to authentic work contexts where AI‑native capabilities can develop.

These challenges demand infrastructure, not programs. Recent research reveals that skills exhibit hierarchical dependency structures, and foundational skills are prerequisite for advanced skills in nested patterns (Hosseinioun et al., 2025). This finding supports the pyramid's functional dependency logic, not as a pathway individuals must ascend, but as a system‑level requirement that higher‑order capabilities depend on the presence of foundational ones. It also underscores a central measurement challenge: assessing advanced capabilities requires first verifying prerequisite competencies through fine‑grained, competency‑based evaluation rather than coarse educational proxies.

Society‑wide capability development therefore requires systems that can continuously update skill definitions, provide accessible pathways for capability acquisition across populations, enable reliable competency‑based verification, and bridge the gap between learning environments and real work contexts.

## Implications for Governments and Workforce Policy

At the national level, governments face a fundamental challenge: they must assess and plan the distribution of AI capabilities across sectors, regions, and population segments without reliable mechanisms to do so. Traditional workforce planning tools, labor force surveys, occupational classifications, educational attainment data, cannot capture the dynamic, contextual nature of AI-related capabilities.

Governments need new infrastructure to make AI capability visible, measurable, and actionable for policy. This requires two foundational systems: **skills ontologies** that map AI-related competencies to occupational roles and learning pathways, and **workforce intelligence systems** that integrate data from training providers, labor market outcomes, and credentialing bodies in real time. These systems require competency‑based measurement frameworks grounded in shared skill ontologies, enabling governments to move beyond educational proxies toward real‑time visibility into AI‑related capabilities across the workforce.

Together, these systems constitute a form of digital public infrastructure for workforce development—not episodic policy programs, but continuous capabilities that enable governments to target investments, detect capability gaps, and design interventions with precision. The AI Pyramid provides the organizing framework: governments can assess whether they possess sufficient AI‑native capability at the population base, adequate Foundation‑level capacity to build and sustain systems, and access to Deep‑level expertise through national or global research ecosystems to maintain technological and economic resilience.

More fundamentally, AI capability infrastructure—spanning measurement, learning, and credentialing—allows governments to address inequality arising from differential access to AI capabilities. Brynjolfsson et al.'s (2025) finding that AI assistance most benefits lower-skilled workers suggests AI could reduce skill-based wage gaps, but only if lower-skilled populations gain access to AI tools and develop competency using them. Without systematic, competency-based visibility into who possesses AI-related skills and how those skills are distributed, inequality remains latent until it manifests as persistent economic divergence.

## Conclusion

The AI Pyramid provides a conceptual framework for understanding how societies can organize human capability in an era where cognition is increasingly mediated by machine systems. By distinguishing between AI Native, Foundation, and Deep capabilities, the framework clarifies not only which competencies must be widespread and which must remain specialized, but also how these capabilities function together as a system rather than as a linear progression. Central to this model is the recognition that effective AI adoption depends on a shared, competency-based capability infrastructure—encompassing measurement, learning, and credentialing—adapted across layers rather than reinvented for each one.

Important research gaps remain. The framework calls for empirical validation across organizational, sectoral, and national contexts: Do societies with more balanced distributions of AI-native, Foundation, and Deep capabilities realize greater productivity, resilience, or inclusion? How do optimal capability distributions vary across industries and stages of economic development? What competency-based assessment instruments can reliably distinguish behavioral fluency, system-building capability, and frontier expertise in dynamic, real-world settings?

As AI continues to diffuse across economic and social domains, the capacity to deliberately measure, cultivate, and distribute human capabilities will become a central determinant of institutional adaptability, economic competitiveness, and social inclusion. The AI Pyramid offers a foundation for this effort, but realizing its potential requires sustained research, policy experimentation, and cross-sector collaboration to translate conceptual clarity into durable capability infrastructure at scale.